\documentclass[12pt, final]{l4dc2023}

\usepackage{graphics} 
\usepackage{amstext}
\usepackage{bm}
\usepackage{mathtools}
\usepackage[capitalise]{cleveref}
\usepackage[font=scriptsize,labelfont=bf,tableposition=top]{caption}
\usepackage{booktabs}
\usepackage{adjustbox}

\SetKwFor{ForPar}{for}{do in parallel}{end forpar}
\definecolor{alg}{RGB}{46, 149, 186}

\SetCommentSty{mycommfont}
\definecolor{urldarkblue}{RGB}{1, 111, 255}

\title[HiPBOT]{Hierarchical Policy Blending As Optimal Transport}

\author{%
 \Name{An T. Le} \Email{an.le@tu-darmstadt.de}\\
 \addr Computer Science Department, TU Darmstadt, Germany
 \AND
 \Name{Kay Hansel} \Email{kay.hansel@tu-darmstadt.de}\\
  \addr Computer Science Department, TU Darmstadt, Germany
 \AND
 \Name{Jan Peters} \Email{jan.peters@tu-darmstadt.de}\\
 \addr Computer Science Department, TU Darmstadt, Germany\\
 \addr DFKI, RD SAIRoL, Hessian.AI, Centre for Cognitive Science 
 \AND
 \Name{Georgia Chalvatzaki} \Email{georgia.chalvatzaki@tu-darmstadt.de}\\
  \addr Computer Science Department, TU Darmstadt, Germany and Hessian.AI
}

\newcommand{\mc}[1]{\mathcal{#1}}

\renewcommand{\vec}[1]{\textbf{\textit{#1}}}

\DeclareMathOperator*{\argmax}{\arg\!\max}
\DeclareMathOperator*{\argmin}{\arg\!\min}

\DeclarePairedDelimiterX{\doubleVertBar}[2]{[}{]}{%
  #1\;\delimsize\|\;#2%
}

\renewcommand{\H}[2][]{
\ifthenelse {\equal{#1}{}}
{\mathbb{H}\left[#2\right]}
{\mathbb{H}_{#1}\left[#2\right]}}

\newcommand{\E}[2][]{
\ifthenelse {\equal{#1}{}}
{\mathbb{E}\left[#2\right]}
{\mathbb{E}_{#1}\left[#2\right]}}

\newcommand{\Sspace}{\mc{S}}
\newcommand{\Aspace}{\mc{A}}

\newcommand{\nexperts}{n}

\newcommand{\temperature}{\beta}

\def\1{\bm{1}}

\def\RR{\mathbb{R}}

\def\eps{{\epsilon}}

\def\vtheta{{\bm{\theta}}}
\def\vpi{{\bm{\pi}}}
\def\va{{\bm{a}}}

\def\vc{{\bm{c}}}

\def\vf{{\bm{f}}}

\def\vm{{\bm{m}}}
\def\vn{{\bm{n}}}

\def\vq{{\bm{q}}}

\def\vs{{\bm{s}}}

\def\vw{{\bm{w}}}
\def\vx{{\bm{x}}}

\def\vz{{\bm{z}}}

\def\vbeta{{\boldsymbol{\beta}}}

\def\vtheta{{\boldsymbol{\theta}}}

\def\mC{{\bm{C}}}

\def\mJ{{\bm{J}}}

\def\mM{{\bm{M}}}

\def\mP{{\bm{P}}}

\DeclareMathAlphabet{\mathsfit}{\encodingdefault}{\sfdefault}{m}{sl}
\SetMathAlphabet{\mathsfit}{bold}{\encodingdefault}{\sfdefault}{bx}{n}

\def\gA{{\mathcal{A}}}

\def\gN{{\mathcal{N}}}

\def\gQ{{\mathcal{Q}}}

\def\gT{{\mathcal{T}}}

\def\gX{{\mathcal{X}}}

\DeclareMathOperator{\defi}{def}
\DeclareMathOperator{\defeq}{\overset{\defi}{=}}

\makeatletter
\newcommand{\StatexIndent}[1][3]{%
  \setlength\@tempdima{\algorithmicindent}%
  \Statex\hskip\dimexpr#1\@tempdima\relax}
\makeatother

\newcommand{\dotprod}[2]{\ensuremath{\langle #1 , #2\,\rangle}}

\def\ones{\bm{1}}

\def\J\mathbf{J}

\DeclarePairedDelimiterX{\infdivx}[2]{(}{)}{%
  #1\;\delimsize\|\;#2%
}

\newcommand{\hide}[1]{}

\begin{document}
\vspace{-0.8cm}
\maketitle
\vspace{-0.6cm}
\begin{abstract}%
We present hierarchical policy blending as optimal transport (HiPBOT). HiPBOT hierarchically adjusts the weights of low-level reactive expert policies of different agents by adding a look-ahead planning layer on the parameter space. The high-level planner renders policy blending as unbalanced optimal transport consolidating the scaling of the underlying Riemannian motion policies. As a result, HiPBOT effectively decides the priorities between expert policies and agents, ensuring the task's success and guaranteeing safety. Experimental results in several application scenarios, from low-dimensional navigation to high-dimensional whole-body control, show the efficacy and efficiency of HiPBOT. Our method outperforms state-of-the-art baselines -- either adopting probabilistic inference or defining a tree structure of experts -- paving the way for new applications of optimal transport to robot control.
More material at \url{https://sites.google.com/view/hipobot}.


\end{abstract}


\section{Introduction}
\begin{figure}[b]
\vspace{-0.3cm}
\centering
  \begin{minipage}[b]{0.49\textwidth}
  \centering
    \includegraphics[width=0.75\linewidth]{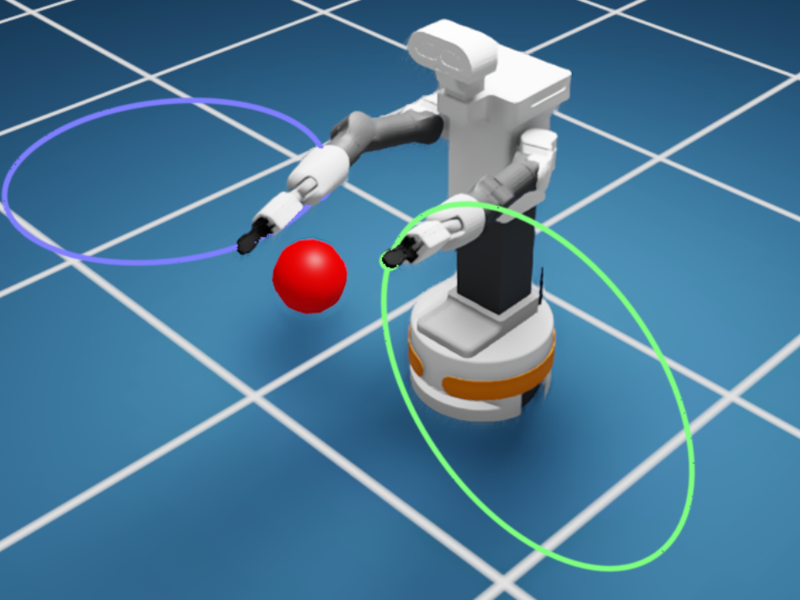}
  \end{minipage}
  \begin{minipage}[b]{0.49\textwidth}
  \centering
    \includegraphics[width=0.75\linewidth]{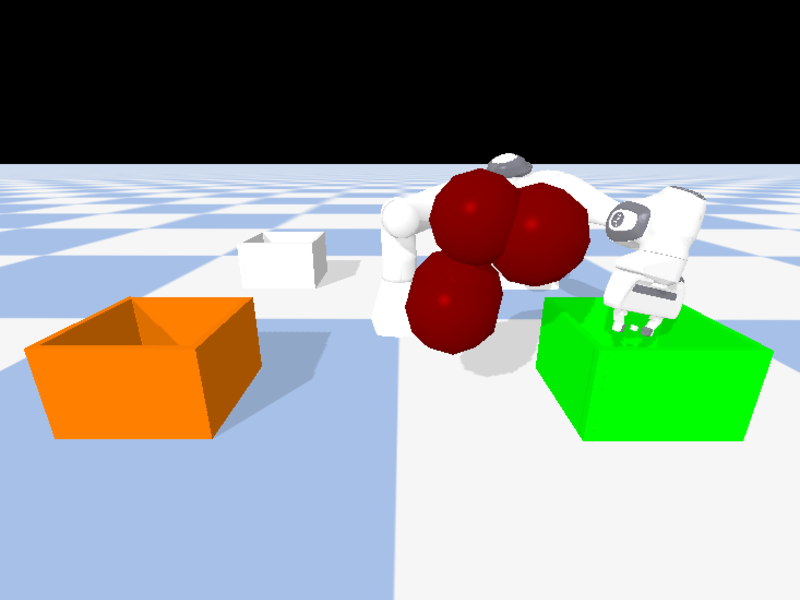}
  \end{minipage}
  \caption{(Left) TIAGo++ whole-body control environment. The two arms of TIAGo++ are treated as two agents tracking different reference trajectories while avoiding collisions. (Right) Panda manipulation environment. The Panda has to start from a randomly selected box (brown) and reach the randomly selected target box (green) while avoiding all possible collisions with itself and the dynamic environment (red obstacles can be static or moving with constant velocity).}
\label{fig:high_dof}
\vspace{-0.6cm}
\end{figure}
Reactive motion generation is a fundamental functionality for robots operating in complex and unstructured settings, where dynamic changes can occur~\citep{park2008movement,hogan2012dynamic,ijspeert2013dynamical,paraschos_2018_promp,ratliff_rmp_2018}. Reactivity emerges through high-frequency control policies that can resolve a specific behavior, e.g., approaching a goal, or avoiding an obstacle~\citep{khatib1987unified} or combinations of such behaviors~\citep{cheng_rmpflow_2018}.

Typical methods in literature achieve such reactivity through three key approaches. First, operational space control (OSC) with a hierarchy of tasks~\citep{khatib1987unified,khatib2004whole}, where a sequence of quadratic programming (QP) problems is resolved in the nullspace of the previous task in the priority list~\citep{flacco2015control}. Second, reactive motion generation through the synthesis of Riemannian Motion Policies (RMPs) defined in a tree structure~\citep{cheng_rmpflow_2018,cheng2021rmpflow}, which generalizes the OSC framework to geometric dynamical systems (GDS)~\citep{bullo2019geometric}. The third approach refers to learning-based approaches that use either demonstrated data -- imitation learning~\citep{ijspeert2013dynamical,paraschos_2018_promp}-- or trial-and-error interactions with the environment -- reinforcement learning (RL)~\citep{kober2013reinforcement, moos2022robust} -- for learning to react and adapt to domain changes.

On the opposite side of the spectrum lie planning-based approaches -- either sampling-based~\citep{kavraki_1996_probabilistic, lavalle_1998_rapidly, lavalle_2006_planning,kalakrishnan_2011_stomp,bhardwaj_storm_2022} or gradient-based~\citep{zucker_2013_chomp, mukadam_2018_gpmp} -- which simulate a look-ahead forward modeling of the environment, towards reaching a predetermined goal. 
The interplay of planning and reactivity is long studied in robotics~\citep{kaelbling2010hierarchical,srivastava2014combined,pertsch2020long,bahl2021hierarchical}, and it frequently emerges when a high-level agent sets subgoals towards a long-term goal for a myopic reactive agent to achieve. This hierarchical planning and control can be realized by explicitly setting task sub-goals~\citep{jauhri2022rlmmbp,xia2020relmogen,sharma2021autonomous}, or by adapting parameters of the low-level policy towards a single long-term goal~\citep{kroemer2015towards,end2017layered,celik2022specializing,akrour2021continuous,zaki_2022_improperRL}.

 Our work focuses on the latter aspect of hierarchical control. We present a planning approach on the parameter space of given reactive policies describing different robot behaviors. Representing the low-level policies in exponential form allows a seamless composition in the form of a product of experts. We frame the look-ahead planning as an Unbalanced Optimal Transport (UOT) problem redistributing the multimodal weights of the expert policies. This formulation enables the simultaneous adaptation of expert policies utilized by multiple agents. Hence, our approach consolidates the low-level experts of various agents and helps in practice to avoid local minima in highly dynamic environments.

To summarize, our contributions are \textbf{(i)} introducing the definition of multi-expert-multi-agent blending problem, \textbf{(ii)} casting the blending problem as entropic-regularized UOT, thereby utilizing its convexity and efficient solver. And, \textbf{(iii)} we show that by realizing experts as RMPs, we can maintain asymptotic local stability.

\section{Preliminaries And Problem Statement}
We introduce preliminaries for defining stable experts for reactive motion generation. Then, we state the policy-blending problem by representing the experts in exponential form and cast it as an entropic-regularized UOT.

\subsection{Riemannian Motion Policy}
An RMP~\citep{ratliff_rmp_2018} is a mathematical object $(\vpi, \mM)$ representing reactive, modular, and composable motion generation policies, where $\vpi$ is a deterministic policy mapping states to actions, and $\mM$ is the Riemannian matrix representing the policy weight. The state ${\vs = (\vq, \dot{\vq}, \vc)}$ represents the robot's position $\vq\in \RR^q$, velocity $\dot{\vq}\in \RR^q$ and environment context $\vc$. We assume a set of homeomorphic task maps $\{\phi_i: \gQ \xrightarrow{} \gX_i\}$, that relate the robot configuration $\gQ$ space and a certain task space $\gX_i$ of the $i^{\text{th}}$ task. Then, given a set of task-space policies $(\vpi^{\gX_i}, \mM^{\gX_i})$, we can represent a deterministic acceleration policy in the robot configuration space by ${\vpi = \va = \ddot{\vq} = \mM^{\dagger} \sum_i\mJ_{\phi_i}^{\intercal} \mM^{\gX_i}\vpi^{\gX_i}(\phi(\vs))}$, with $\mM = \sum_i \mJ_{\phi_i}^{\intercal} \mM^{\gX_i} \mJ_{\phi_i}$, $\mJ_{\phi_i}$ the Jacobian of the task map $\phi_i$ and $\dagger$ is the pseudo-inverse operator.
\vspace{-0.3cm}
\subsection{Product of Experts}

To define the policy blending problem, we formalize each expert policy $i \in \{1, \ldots, n\}$ in a set of $n$ policies as the Boltzmann distribution form
$
    \pi_{i}(\va \mid \vs; \vtheta_{i})
    \propto
    \exp(-E_{i}(\vs, \va; \vtheta_{i})),
$
where the quantities $\vs \in \Sspace$ and $\va \in \Aspace$ denote a state and an action, respectively. An energy function $E_{i}:\Sspace \times \Aspace \rightarrow \RR$ assigns a cost to each state-action pair. The choice of the energy function $E_{i}$ and its hyperparameter $\vtheta_{i}$ is usually designed or learned in advance. Following the PoE~\citep{hinton2002training} formulation, the blended policy for an agent can be defined as
\begin{equation}
    \label{eq:poe}
    \pi(\va \mid \vs,\;\vbeta) = \prod_{i=1}^{\nexperts} \pi_{i}(\va \mid \vs;\;\vtheta_{i})^{\temperature_i} \propto \exp\left(-\sum_{i=1}^{\nexperts} \temperature_i  E_{i}(\vs, \va; \vtheta_{i}) \right)
\end{equation}
with blending/weighting factors $\vbeta$, also known as \textit{temperatures}, representing the importance or relevance of each policy in the product. In the logarithmic space, this policy blending corresponds to a weighted superposition by performing Maximum Likelihood (MLE) on the PoE 
$
    \va^* = \argmax_{ \va \in \Aspace } \log \pi (\va \mid \vs, \vbeta) = \argmin_{ \va \in \Aspace } \sum_{i=1}^{\nexperts} \temperature_i  E_{i}(\vs, \va; \vtheta_{i})
$
depending on state $\vs$ and $\vbeta$. We can further formulate the state-dependent temperature $\vbeta(\vs)$, giving the possibility to change the state-dependent relevance or importance of experts. In an online fashion, a change in the expert weighting (i.e., expert relevancy) makes it possible to induce \textit{parameter planning} into the myopic nature of the policy $\pi( \va \mid \vs)$. In particular, the simplest form of such parameter planning problem (i.e., a policy blending problem in our paper) can be formulated as the linear program
$
 \min_{\vbeta \in \RR_+^{n}} \dotprod{\vbeta}{\mC(\vs)},
$
with $\langle \cdot,\cdot \rangle$ is the Frobenius dot-product, and a state-dependent objective matrix $\mC(\vs)$ (e.g., rollout return in RL settings) dictating the situational blending weights $\vbeta$. We realize the PoE in the RMP framework utilizing its stability property, as shown in next sections.
\vspace{-0.3cm}
\subsection{Preliminaries On Optimal Transport}
We give a brief introduction to OT and motivate why solving the linear program of policy blending using OT.

\noindent\textbf{Histograms And Transport Polytope.}~For two histograms $\vn \in \Sigma_n$ and $\vm \in \Sigma_m$ in the simplex $\Sigma_d \defeq \{\vx \in \RR^d_+: \vx^\intercal\ones_d=1\}$, we define the transport polytope $U(\vn, \vm)$, namely the set of $n \times m$ matrices
$
  U(\vn, \vm) \defeq  \{\mP \in \RR_+^{n \times m}\; |\; \mP\ones_m=\vn, \mP^\intercal\ones_n=\vm\} \, 
$
where $\ones_d$ is the $d$-dimensional vector of ones. From a probabilistic view, the set of $U(\vn, \vm)$ contains all possible \emph{joint probabilities} of two multinomial random variables $(X, Y)$ having histograms $\vn$ and $\vm$, respectively. Indeed, any matrix $\mP \in U(\vn, \vm)$ can be identified as the joint probability for $(X, Y)$ such that $p(X=i,Y=j)=p_{ij}$. We define the entropy $H(\cdot)$ of these histograms and their marginals as
$
\vx \in \Sigma_n,\, H(\vx)=-\sum_{i=1}^n {x_i} \log {x_i},\, H(\mP)=-\sum_{i,j=1}^{n, m} p_{ij} (\log p_{ij} - 1).
$

\noindent\textbf{Entropic-Regularized Optimal Transport.}~Given a $n \times m$ cost matrix $\mC$, the OT between $\vn$ and $\vm$ given cost $\mC$ is
$
    d_C(\vn, \vm) \defeq \min_{\mP\in U(\vn, \vm)} \dotprod{\mP}{\mC},
$
which is exactly the above policy blending problem with additional histogram constraints. However, solving this linear program is expensive for large matrix dimensions in our reactive motion generation setting. For a general matrix $\mC$ and $d = \max (n, m)$, the worst case complexity of classical algorithms~\citep{ahuja1988network, orlin1988faster} solving for the optimal plan $\mP^*$ is $O(d^3\log d)$. To deal with the scalability of computing the OT,~\citet{cuturi2013sinkhorn} propose to regularize its objective function by the entropy of the transport plan, which results in the entropic-regularized OT with an entropy scaling scalar $\lambda$
\begin{equation}\label{eq:sinkhorn_dist}
\mP^\lambda \defeq \argmin_{\mP\in U(\vn, \vm)} \dotprod{\mP}{\mC} - \lambda H(\mP).
\end{equation}
The solution $\mP^\lambda$ is unique due to the strict convexity of the negative entropy term. The entropic regularization enables the celebrated Sinkhorn-Knopp algorithm~\citep{sinkhorn1967diagonal} to solve OT, shown to have a complexity of $\Tilde{O}(d^2/\epsilon^3)$~\citep{altschuler2017near}, with $\eps$ is the relaxation error.

\noindent\textbf{Entropic-Regularized Unbalanced Optimal Transport.}~A key constraint of classical OT is that it requires the input measures to be normalized to the unit mass, which is a problematic assumption for many applications that require handling arbitrary positive measures (mass creation or destruction), and/or allowing for only partial displacement of mass. The entropic-regularized UOT~\citep{frogner2015learning, chizat2018scaling} is defined as
\begin{equation}\label{eq:uot}
\mP^{\lambda}_{UOT}  \defeq \argmin_{\mP\in \RR_+^{n \times m}}\dotprod{\mP}{\mC} - \lambda H(\mP) + \lambda_{KL} \left(\widetilde{\mathrm{KL}}(\mP\ones_m \;\|\; \vn) + \widetilde{\mathrm{KL}}(\mP^\intercal\ones_n \;\|\; \vm)\right)
\end{equation}
where now $\vn \in \RR_+^n, \vm \in \RR_+^m$ are arbitrary positive vectors, $\lambda_{KL}$ is the marginal regularization scalar, and $\widetilde{\mathrm{KL}}(\vw || \vz) = \vw^\intercal\log (\vw \oslash \vz) - \ones^\intercal\vw + \ones^\intercal\vz$ is the generalized Kullback-Leibler (KL) divergence between two positive vectors $\vw, \vz \in \RR_+^k$ ($\oslash$ is the element-wise division), with the convention $0 \log 0 = 0$.~\citet{pham2020unbalanced} show that the (Sinkhorn-like) generalized matrix scaling algorithm~\citep{frogner2015learning} solves the dual of (\ref{eq:uot}) with the complexity of $\Tilde{O}(d^2/\epsilon)$ and is guaranteed to converge (Theorem 4.1 in~\citep{chizat2018scaling}). With these properties, casting policy blending as an entropic-regularized UOT problem is desirable since UOT relaxes the normalizing constraint, bringing the OT problem back to the linear program form of policy blending, as policy blending weights are usually unnormalized quantities. Moreover, assuming a Gaussian distribution over blending weights is often non-realistic, as multiple policies may have similar priorities given the current situation. Finally, the entropic-regularized UOT benefits from the computation efficiency of the Sinkhorn-like algorithm.

\section{Hierarchical Policy Blending As Optimal Transport}\label{sec:pbot}
In this section, we propose \textbf{Hi}erarchical \textbf{P}olicy \textbf{B}lending as \textbf{O}ptimal \textbf{T}ransport (HiBPOT) - a two-level hierarchical scheme for reactive motion generation. We hypothesize there exist multiple agents controlling the same dynamical system satisfying some objectives at the upper-level, where each agent utilizes myopic (learned or crafted) expert policies to compute their actions at the lower-level~\citep{hansel2022hierarchical}. In particular, each agent can choose to control a subset or all DoFs of the dynamical system. The upper level employs entropic-regularized UOT to solve the policy blending problem by observing expert rollouts that inform the weight-scaling of the lower-level agents.

\subsection{Product Of Experts-Agents}

Let us consider multi-arm systems (Fig. \ref{fig:high_dof}-left), where each robotic arm can be considered an agent acting on the whole system to execute some tasks. Assuming the agents' behaviors are collaborative and that there exists a pool of $n$ experts and $m$ agents, we propose a simple solution for the Multi-Experts-Multi-Agents (MEMA) policy blending problem by extending the PoE \eqref{eq:poe}, as defined in Definition \ref{def:mema}.
\begin{definition}[Product of Experts-Agents] \label{def:mema}
Let $\vbeta \in \RR_+^{n \times m}$ be the positive blending weight matrix for $n$ experts and $m$ agents. The MEMA blending policy is defined as the product of experts-agents (PoEA) 
\begin{equation}\label{eq:mema}
\pi(\va \mid \vs,\;\vbeta) = \prod_{i, j=1}^{n, m} \pi(\va_{i, j} \mid \vs;\;\vtheta_{i, j})^{\temperature_{i, j}} \propto \exp\left(-\sum_{i, j=1}^{n, m} \temperature_{i, j}  E(\vs, \va_{i, j};\;\vtheta_{i, j}) \right)
\end{equation}
with $i,j$ index the $i^{\text{th}}$-expert and $j^{\text{th}}$-agent, $\vs$ is the holistic system state observed by all experts, and $\va$ is the blended pullbacked action from all experts and agents.
\end{definition}
We realize the lower-level experts within the RMP framework. In RMP, for the $i^{\text{th}}$-expert of $j^{\text{th}}$-agent, the task-space energy $E(\vx, \va_{i,j}; \vtheta_{i,j})$ is usually designed as a quadratic function having smooth and convex properties in the task space $\vx \in \gX_{i,j},\,\va_{i,j} \in \gA_{i,j}$, with corresponding task map $\vx = \phi_{i,j}(\vs)$. Accordingly, the Boltzmann distribution forms a Gaussian $\pi(\va_{i,j} \mid \vx; \vtheta_{i,j}) = \gN(\mM_{i,j}(\vx)^{-1}\vf_{i,j}(\vx), \mM_{i,j}(\vx)^{-1})$
locally at $\vx$ with the forcing function $\vf_{i,j}(\vx)$ and $\mM_{i,j}(\vx)$ as the mean and the Riemannian matrix (i.e., the precision matrix), respectively. Within the PoEA view, the \textit{pullbacked} forcing term and Riemannian matrix of $j^{\text{th}}$-agent's configuration policy would be 
$
\vf_j(\vs) = \sum_{i=1}^n \beta_{i,j}(\vs) \mJ_{\phi_{i,j}}^{\intercal} \vf_{i,j}(\vx), \,\mM_j = \sum_{i=1}^n \beta_{i,j}(\vs) \mJ_{\phi_{i,j}}^{\intercal}  \mM_{i,j}(\vx) \mJ_{\phi_{i,j}}
$, respectively. Given the current state and determined temperatures, the MLE blended action (at configuration space) can be computed analytically in closed form~\citep{ratliff_rmp_2018} as
\begin{equation}\label{eq:agent_optimal_a}
\va^* = \argmin_{ \va \in \Aspace } \sum_{i, j=1}^{n, m} \temperature_{i, j}  E(\vs, \va_{i, j};\;\vtheta_{i, j}) = \sum_j \mM_j^\dagger \vf_j(\vs).
\end{equation}

\subsection{Policy Blending As Entropic-Regularized Unbalanced Optimal Transport}

In practice, we found that the normalizing constraint of the policy temperature is restrictive, as it requires spreading enough masses to the policy weight-scaling, leading to overestimation. On the other hand, in case of large number of experts, the normalized temperature matrix puts too small masses on expert policies, thus leading to underestimation and making them underperforming. Thus, we propose to cast policy blending as an entropic-regularized UOT problem to relax the need for the normalizing constraint.

\begin{definition}[MEMA Policy Blending] \label{def:ot_problem}
Let $\vn \in \RR_+^n, \vm \in \RR_+^m$ be arbitrary positive vectors representing the priors of expert-policy and agent-policy temperatures, respectively. The entropic-regularized UOT for the policy blending is defined as
\begin{equation}\label{eq:policy_blending_ot}
\vbeta^*(\vs) = \argmin_{\vbeta \in \RR_+^{n \times m}}\dotprod{\vbeta}{\mC} - \lambda H(\vbeta) + \lambda_{KL} \left(\widetilde{\mathrm{KL}}(\vbeta\ones_m \;\|\; \vn) + \widetilde{\mathrm{KL}}(\vbeta^\intercal\ones_n \;\|\; \vm)\right)
\end{equation}
with $\mC(\vs)$ is the state-dependent cost matrix, which can be learned or computed analytically.
\end{definition}
The solution of \eqref{eq:policy_blending_ot} is unique due to the strict convexity of the objective in $\vbeta$. Due to the uniqueness of the solution and the practical computation complexity of the Sinkhorn-like algorithm solving (\ref{eq:policy_blending_ot}), it is well-suited for reactive motion generation. Note that this formulation optimizes the blending temperatures depending on the objective costs at the upper-level, while still assuming expert independency at the lower-level.



In the dynamics settings of motion planning, the objectives are usually goal-reaching, obstacle avoidance, and self-collision avoidance in dynamic settings. Hence, we follow these objectives to design the state-dependent cost matrix as
\begin{equation}\label{eq:cost_matrix}
\left[\mC(\vs)\right]_{i, j} = \frac{1}{h}\sum_{t=1}^h w_g\underbrace{d(\vs^t_{i, j}, \vs^g_j)}_{\textrm{Goal Cost}} + w_c \underbrace{\exp\left(-\frac{\textrm{SDF}_j(\vs^t_{i, j})^2}{2l^2}\right)}_{\textrm{Collision-Avoidance Cost}}
\end{equation}
where from the current system state $\vs$, the rollout with horizon $h$ from the perspective of $j^{\text{th}}$-agent following the $i^{\text{th}}$-expert is $\{\vs, \vs_{i, j}^1, \ldots, \vs^h_{i, j}\}$. We assume the transition dynamics $\gT(\cdot, \cdot)$ of the system known, and an expert rollout is computed by following $\vs^{t+1} = \gT(\vs^t, \va^*),\, \va^* = \argmin E(\vs, \va; \vtheta)$. $\vs^g_j$ is the $j^{\text{th}}$-agent goal state, and $\textrm{SDF}_j(\cdot)$ is the signed distance field measuring the closest distance of the $j^{\text{th}}$-agent's robot links to obstacles including itself (i.e., self-distance). $l$ is the hyperparameter for the collision margin. Note that the goal or obstacles can be changed over time; thus, the goal and distance field are also updated in the loop. This cost design enables integration of additional higher levels of planning abstractions, e.g., task planning, where the symbolic planner can set the intermediate goals or other contexts in the cost matrix (but this is not integrated in the current work). Since experts are independent by assumption at the lower-level, the elements of the cost matrix can be computed in parallel using GPU~\footnote{An algorithm overview of our HiPBOT framework can be found on the project website.}. 


\subsection{Stability Analysis}

HiPBOT, as we deploy as expert policies RMPs, only sets the scaling factor for both $\vf_{i, j}, \mM_{i, j}$ at the lower-level. Analyzing its local stability is straightforward.
\begin{proposition}[Asymptotic Stability]
As $\vbeta \in \RR_+^{n \times m}$ by Definition \ref{def:ot_problem} is positive, if all expert RMPs are in the form of Geometric Dynamical Systems, then by \textbf{Theorem 2} in~\citep{cheng_rmpflow_2018}, the system that follows the HiPBOT policy as Product of Experts-Agents in Definition \ref{def:mema} converges to the forward invariance set $\mathcal{C}_{\infty}:=\left\{(\vs, \dot{\vs}): \vf(\vs)=0, \dot{\vs}=0\right\}$.
\end{proposition}
Note that this local stability of HiPBOT is only valid for static environments, where the parameters for collision-avoidance RMPs do not change. Nevertheless,
for dynamic environments, in most cases, we empirically observed that the agents also exhibit locally stable behaviors, and we plan to investigate theoretically further in the future.

\section{Experiments}\label{sec:experiments}
We evaluate HiPBOT in two toy environments and a manipulation task with a 7DoF robot. These settings are multi-experts-one-agent. Finally, we demonstrate HiPBOT in whole-body reactive motion generation.

\noindent\textbf{Baselines.} We benchmark HiPBOT against two baselines. First, RMPflow~\citep{cheng_rmpflow_2018}, a myopic baseline without look-ahead evaluation, composes all expert RMPs to generate a global dynamical behavior. It also runs at a very high frequency due to low computational demands. Second, as a strong baseline, HiPBI~\citep{hansel2022hierarchical} is a similar hierarchical scheme to our HiPBOT, which adopts probabilistic inference to address the blending problem. HiPBI samples the rollouts from a temperature proposal distribution and updates the temperature distribution in an online fashion, while HiPBOT shoots individual experts evaluating their contributions assuming unnormalized temperature priors. For both HiPBOT and HiPBI, the environment and the methods operate asynchronously. The algorithms must make quick decisions to react to unexpected changes in the landscape.

\noindent\textbf{Metrics and Settings}. We use the following metrics: (i) success rate (SUC), indicating the percentage of goal reaching without any collisions; (ii) safety rate (SAFE) of collision-free motions regardless of reaching the goal or not; (iii) the final l2 distance (l2D) to the goal; and (iv) the total time steps (TS) need until the goal is reached. For a comparative study, we use different rollout horizons for HiPBI and HiPBOT.


\subsection{Toy Environments}
\begin{figure}[tb]
\centering
  \begin{minipage}[b]{0.4\textwidth}
    \includegraphics[width=\linewidth]{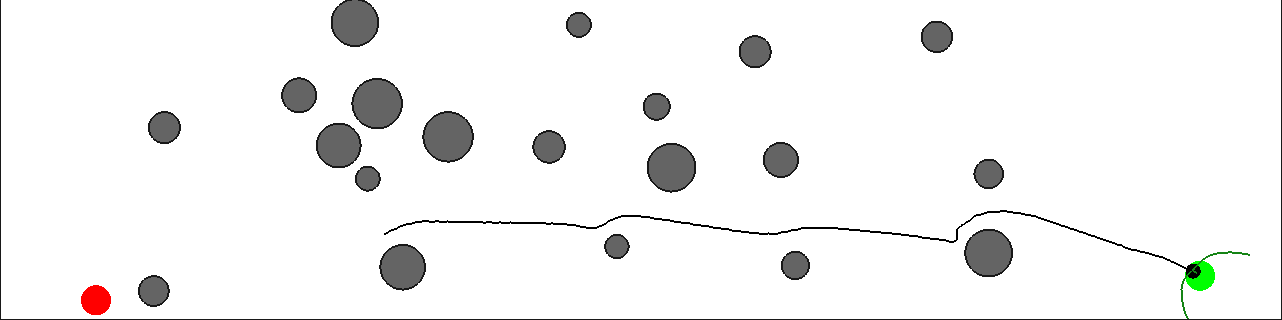}
  \end{minipage}
    \begin{minipage}[b]{0.4\textwidth}
    \includegraphics[width=\linewidth]{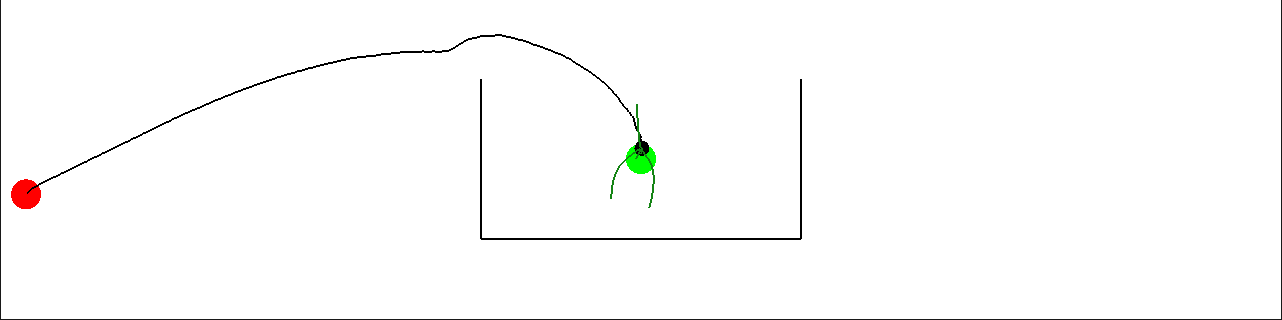}
  \end{minipage}
  \caption{(Left) Planar Maze Environment. We randomly sample $K$ circular obstacles inside a restricted area between the start and goal positions. We model the movement of the obstacles using simple Euler integration. (Right) Planar Box Environment. The box can be either static or dynamic, and its motion is modeled as a constant velocity. In both environments, an agent moves from a random start (red) to a random goal (green) position. Green lines are expert rollouts.}
\label{fig:toy_env}
\vspace{-0.6cm}
\end{figure}

The \textbf{Planar Maze Environment} is a cluttered environment, see Fig.~\ref{fig:toy_env}. This maze environment mimics a dense, cluttered, and dynamic environment. In this case, local minima are created but often disappear independently. However, the control methods have to be reactive enough to avoid collisions. Unlike the maze, the \textbf{Planar Box Environment} is a sparse domain. The agent start position is sampled randomly to the right or left of the box. The challenge lies in not getting into a local optimum in front (left or right) or below the box. Furthermore, the dynamic nature complicates the planning of a promising solution. In this case, although the reactive requirement is relaxed due to being sparse, the difficult local minima always exist.
We design common RMPs such as collision avoidance and goal reaching for all methods, as in~\citep{cheng_rmpflow_2018}. To achieve a curving behavior, we design an expert $\pi_{\mathbf{curl}}$ that exerts forces on the normal space of the potential forces and add two opposing curling experts for balancing. Although these curving experts do not affect RMPflow, hierarchical methods achieve curving behavior by adapting blending weights.

\begin{figure}[tb]
\begin{center}
{\scriptsize
\captionof{table}{{Evaluation of HiPBOT versus baselines on the static planar environments. This experiment shows the capabilities of HiPBOT to overcome local minima (100 seeds per evaluation).}}\label{tab:static_toy}
\vspace{-0.25cm}
\tabcolsep1.pt 
\adjustbox{max width=0.89\textwidth}{
\begin{tabular*}{\linewidth}{l
@{\extracolsep{\fill}}
c c c c @{\extracolsep{\fill}} c
c c c c}
\toprule
\phantom{Var.} &  
\multicolumn{4}{c}{2D Toy Box Environment} && \multicolumn{4}{c}{2D Toy Maze Environment}\\
\cmidrule{2-5}
\cmidrule{7-10} 
& {SUC$[\%]$} & {SAFE$[\%]$} & {D2G} & {TS} & {} & {SUC$[\%]$} & {SAFE$[\%]$} & {D2G} & {TS} \\
\midrule
RMP\textit{flow} 
& $0$ & $\mathbf{100}$ & $198.8\,\pm0.7$ & $500.0\,\pm0.0$ 
&
& $73$ & $99$ & $161.5\,\pm296.7$ & $338.7\,\pm100.8$ \\[1ex]

HiPBI ($h=25$)
& $0$ & $\mathbf{100}$ & $198.9\,\pm0.5$ & $500.0\,\pm0.0$ 
&
& $77$ & $93$ & $148.2\,\pm284.5$ & $331.3\,\pm98.9$ \\
HiPBI ($h=5$) 
& $64$ & $\mathbf{100}$ & $82.1\,\pm79.9$ & $354.3\,\pm169.7$ 
&
& $77$ & $97$ & $151.1\,\pm284.1$ & $\mathbf{323.2\,\pm99.3}$ \\
\textbf{HiPBOT} ($h=5$) 
& $0$ & $\mathbf{100}$ & $176.2\,\pm1.2$ & $500.0\,\pm0.0$ 
&
& $72$ & $96$  & $200.0\,\pm334.2$ & $386.2\,\pm234.2$\\
\textbf{HiPBOT} ($h=10$) 
& $\mathbf{93}$ & $\mathbf{100}$ & $\mathbf{17.2\,\pm6.7}$ & $\mathbf{132.9\,\pm10.1}$
&
& $\mathbf{82}$ & $\mathbf{100}$ & $\mathbf{138.5\,\pm300.6}$ & $401.0\,\pm281.6$ \\
\bottomrule
\end{tabular*}}}%
\end{center}
\vspace{-0.7cm}
\begin{center}
{\scriptsize
\captionof{table}{{Evaluation of HiPBOT vs. baselines on the dynamic planar environments with 10-pixel velocity levels. We also compare HiPBI and HiPBOT in synchronous (S) and asynchronous (A) settings. In (S), the simulation waits for the blending solution before the agent steps in the environment. In (A), the environment and the methods act asynchronously. This experiment demonstrates the computational advantage of HiPBOT with short look-ahead horizon, balancing between being reactive and explorative for safety (100 seeds per evaluation).}}\label{tab:dynamic_toy}
\vspace{-0.25cm}
\tabcolsep1.pt 
\adjustbox{max width=0.89\textwidth}{
\begin{tabular*}{\linewidth}{l
@{\extracolsep{\fill}}
c c c c @{\extracolsep{\fill}} c
c c c c}
\toprule
\phantom{Var.} &  
\multicolumn{4}{c}{2D Toy Box Environment} && \multicolumn{4}{c}{2D Toy Maze Environment}\\
\cmidrule{2-5}
\cmidrule{7-10} 
& {SUC$[\%]$} & {SAFE$[\%]$} & {D2G} & {TS} & {} & {SUC$[\%]$} & {SAFE$[\%]$} & {D2G} & {TS} \\
\midrule
RMP\textit{flow} 
& $0$ & $\mathbf{100}$ & $198.9\,\pm1.5$ & $500.0\,\pm0.0$ 
&
& $77$ & $89$ & $161.5\,\pm620.0$ & $330.7\,\pm191.3$ \\[1ex]
HiPBI ($h=25$, S) 
& $2$ & $\mathbf{100}$ & $189.3\,\pm44.7$ & $490.9\,\pm81.8$ 
&
& $98$ & $\mathbf{99}$ & $20.3\,\pm172.7$ & $247.6\,\pm55.8$ \\
HiPBI ($h=5$0, S) 
& $61$ & $\mathbf{100}$ & $49.5\,\pm75.6$ & $276.6\,\pm251.2$
&
& $\mathbf{99}$ & $\mathbf{99}$ & $17.5\,\pm162.6$ & $\mathbf{247.5\,\pm47.6}$ \\
HiPBI ($h=75$, S) 
& $\mathbf{100}$ & $\mathbf{100}$ & $\mathbf{7.3\,\pm5.9}$ & $131.9\,\pm18.0$ 
&
& $\mathbf{99}$ & $\mathbf{99}$ & $\mathbf{19.0\,\pm171.7}$ & $252.1\,\pm47.3$ \\
\textbf{HiPBOT} ($h=5$, S) 
& $0$ & $\mathbf{100}$ & $199.3\,\pm1.1$ & $500.0\,\pm0.0$ 
&
& $\mathbf{99}$ & $\mathbf{99}$ & $26.1\,\pm108.8$ & $315.9\,\pm129.5$  \\
\textbf{HiPBOT} ($h=10$, S) 
& $\mathbf{100}$ & $\mathbf{100}$ & $25.8\,\pm0.6$ & $143.9\,\pm22.8$
&
& $\mathbf{99}$ & $\mathbf{99}$ & $22.0\,\pm72.9$ & $294.2\,\pm108.1$ \\
\textbf{HiPBOT} ($h=15$, S) 
& $\mathbf{100}$ & $\mathbf{100}$ & $16.4\,\pm4.1$ & $\mathbf{127.3\,\pm18.2}$
&
& 98 & 98 & $30.8\,\pm104.5$ & $312.9\,\pm145.6$ \\[1ex]
HiPBI ($h=25$, A) 
& $7$ & $\mathbf{100}$ & $178.6\,\pm71.1$ & $477.1\,\pm120.3$ 
&
& $83$ & $84$ & $116.2\,\pm386.3$ & $294.2\,\pm131.4$ \\
HiPBI ($h=5$0, A) 
& $73$ & $\mathbf{100}$ & $40.1\,\pm76.9$ & $324.3\,\pm169.7$ 
&
& $85$ & $87$ & $100.0\,\pm357.9$ & $\mathbf{293.4\,\pm123.7}$ \\
HiPBI ($h=75$, A) 
& $\mathbf{100}$ & $\mathbf{100}$ & $\mathbf{8.5\,\pm6.0}$ & $205.8\,\pm35.3$
&
& $86$ & $87$ & $106.1\,\pm376.5$ & $297.3\,\pm122.1$ \\
\textbf{HiPBOT} ($h=5$, A) 
& $0$ & $\mathbf{100}$ & $199.1\,\pm1.1$ & $500.0\,\pm0.0$ 
&
& $94$ & $94$  & $55.1\,\pm170.5$ & $321.9\,\pm176.2$\\
\textbf{HiPBOT} ($h=10$, A) 
& $\mathbf{100}$ & $\mathbf{100}$ & $22.2\,\pm4.2$ & $147.9\,\pm20.3$
&
& $\mathbf{99}$ & $\mathbf{99}$ & $\mathbf{20.5\,\pm99.6}$ & $286.0\,\pm80.9$ \\
\textbf{HiPBOT} ($h=15$, A) 
& $\mathbf{100}$ & $\mathbf{100}$ & $17.5\,\pm3.3$ & $\mathbf{126.4\,\pm18.2}$
&
& $94$ & $94$ & $59.8\,\pm203.4$ & $330.6\,\pm188.3$ \\
\bottomrule
\end{tabular*}}}%
\end{center}
\vspace{-0.25cm}
    \centering
    \includegraphics[width=0.9\linewidth]{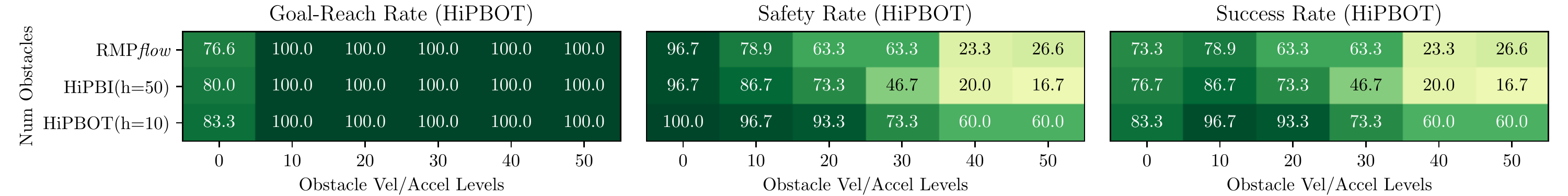}
    \vspace{-0.25cm}
\captionof{figure}{\scriptsize Comparative evaluation for different velocity and acceleration levels of obstacles (30 seeds per evaluation).}
\label{fig:stress_method}
\vspace{-0.4cm}
\end{figure}

\noindent\textbf{Comparative Evaluation.}~Table~\ref{tab:static_toy} shows the comparative results for the static versions of the toy environments. It is evident that the myopic RMPflow is not able to solve the Box domain, but it guarantees safety due to its stable property. Short horizons in HiPBI and HiPBOT are not as effective as longer ones. It is notable that HiPBOT, with only $10$ steps look-ahead, outperforms the baselines in most metrics, guaranteeing maximum safety and good success rates. Table~\ref{tab:dynamic_toy}, shows the dynamic versions of the toy tasks for both synchronous (S) and asynchronous (A) execution of policy blending. HiPBOT outperforms the baselines in all cases with short horizons, making it much faster for deployment in highly dynamic domains. With comparative performance, considering the rollout computation and optimization, we observe that HiPBOT ($h=10$) achieves mean planning rate of around 30Hz due to being efficient with shorter horizon, while HiPBI ($h=50$) runs at about 2Hz. This computing gap hurts the performance of baselines even more in highly dynamic environments, as seen in \cref{fig:stress_method}, depicting a \textit{stress-test} on the 2D Maze for changing acceleration levels of the obstacles by adding Brownian noises. We evaluate a plain goal-reaching success rate--regardless of collisions, along with the safety rate and collision-free success rate. Evidently, HiPBOT with $h=10$ performs overall better, even in the extreme scenarios~\footnote{See videos in our project website \url{https://sites.google.com/view/hipobot/experiment-videos}}.

\begin{figure}[tb]
    \centering
    \vspace{-0.1cm}
    \includegraphics[width=0.89\linewidth]{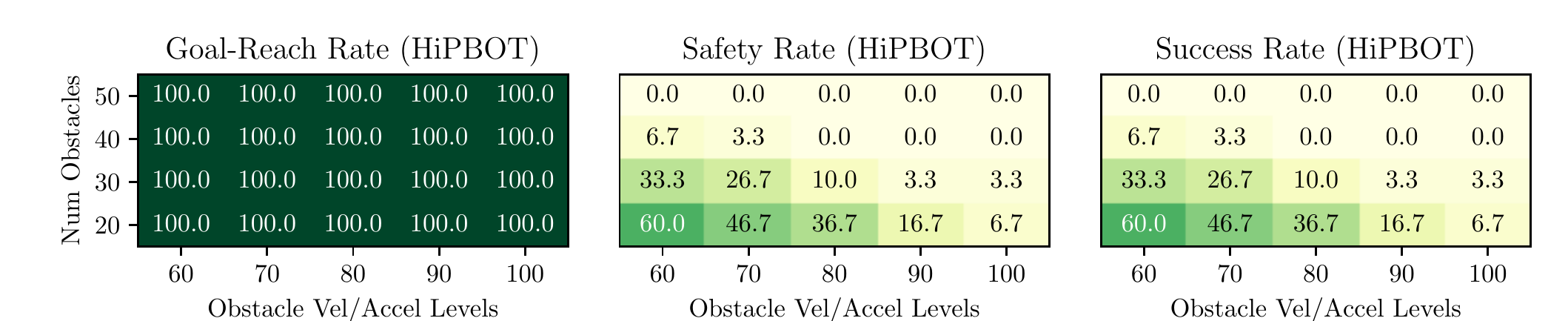}
    \vspace{-0.25cm}
    \caption{Stress test of HiPBOT on extreme velocity and noisy acceleration levels. We run 30 seeds per evaluation.}
    \label{fig:stress_hipbot}
    \centering
    \vspace{-0.125cm}
    \includegraphics[width=0.75\linewidth]{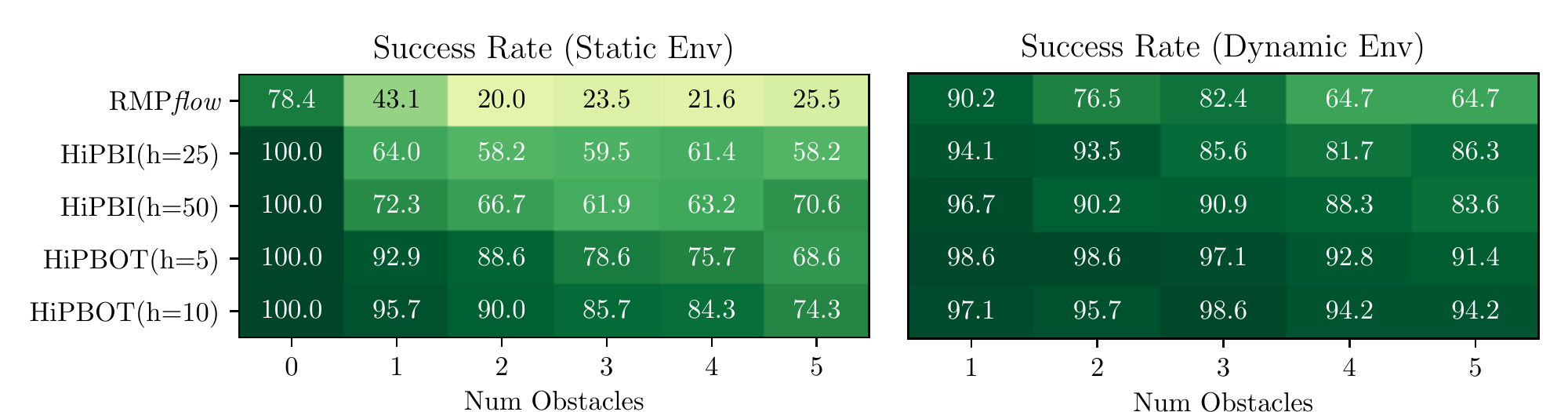}
    \vspace{-0.25cm}
    \caption{Comparative study in the manipulation environment in static and dynamic obstacles, with increasing obstacle number. We run 70 seeds per evaluation.}
    \label{fig:eval_panda}
    \vspace{-0.6cm}
\end{figure}

\noindent\textbf{Stress-test of HiPBOT.}~We even tested more difficult scenarios for the HiPBOT ($h=10$) in the 2D Maze environment, where we varied both numbers of dynamic obstacles and their acceleration levels. While there is total success regardless of collisions, safety is compromised in these extreme dynamic cases.
We hypothesize that longer horizons, more efficient optimization, and even a ``cleverer'' exploration strategy during planning are necessary for these more complex environments.

\subsection{Manipulator Environment}~We test the scalability of our method in high-dimensional manipulation tasks, Fig.~\ref{fig:high_dof}. We implemented eight expert RMPs, ranging from joint \& velocity limits, self-collision to reaching the target, and obstacle avoidance. Both HiPBI and HiPBOT use the same experts with four additional local curling experts for the end-effector that operates in the normal space of the target-reaching potential. While RMPflow cancels out the curling, hierarchical methods adapt the temperatures to achieve the desired dynamic behavior. HiPBOT with horizon 10 is again superior in terms of collision-free success rate. We see that performance drops in the static environment due to difficult local minima that do not vanish over time. This case would require longer look-ahead horizons, but we want to point to the increased efficiency with as few as 10 steps with mean planning rate of 6.4Hz, compared to the 50 steps of HiPBI with mean planning time of 3.5s, which yields lower performance. 

\subsection{Whole-Body Environment}~Finally, we demonstrate HiPBOT capabilities in the MEMA setting with a high-dimensional, multi-objective and highly dynamic environment, see Fig.~\ref{fig:high_dof}. As demonstrated in the videos, HiBPOT is able to compromise between objectives thanks to its ability to adapt expert priorities online. In contrast, RMPflow struggles to find good situational actions and eventually collides.

\vspace{-0.2cm}
\section{Related work}
\noindent\textbf{Reactive motion generation.}~Groundbreaking work was realized with OSC by~\citet{khatib1987unified}, that introduced artificial potential fields for modeling obstacle avoidance (repulsive) and goal-reaching (attractive) behaviors. RMPs~\citep{ratliff_rmp_2018,cheng_rmpflow_2018, xie_gf_2020} consider geodesics in the vicinity of obstacles using Riemannian metrics. Learning-based methods learn reactive and stable primitives~\citep{khansari-zadeh_2011_seds,ijspeert2013dynamical,calinon2014task}. Blending of primitives were introduced in~\citep{luksch2012adaptive,saveriano2019merging}, in probabilistic settings~\citep{paraschos_2018_promp}, and in QP-optimization~\citep{jaquier2022learning}. Recently,~\citet{hansel2022hierarchical} proposed a blending as inference approach for adjusting the weights of a product of experts. Blending also emerges from cost-function formulations as energy-based models~\citep{lambert2022learning,urain2022se}. Fast obstacle avoidance relies on fast perception~\citep{huber2022fast}, and can be realized via safe learning~\citep{liu2022regularized,liu2022safe}. While object-centric primitives are locally reactive, they tend to get stuck in local minima, as they lack look-ahead capabilities. 

\noindent\textbf{Hierarchical planning and control.}~ These approaches refer to multi-level planners or operate in the parameter space of motion policies. The former, such as task and motion planning (TAMP), hierarchical planning, or hierarchical reinforcement learning (HRL)~\citep{kaelbling2010hierarchical,srivastava2014combined,pertsch2020long}, generate sub-goals that an underlying planner or policy must reach. Methods in HRL either adjust constraint functions of dynamic motion primitives~\citep{bahl2020neural,bahl2021hierarchical} or select a policy from a mixture of experts~\citep{daniel2012hierarchical,end2017layered,akrour2021continuous,zaki_2022_improperRL}. Hierarchical mixture of experts selects only one of the experts~\citep{celik2022specializing} to act in the environment. In the case of unexpected environmental changes, this selective behavior leads to sub-optimal performance~\citep{kroemer2015towards}.
We compose simple and stable reactive policies that lead to complex reactive robot behaviors.
\\
\noindent\textbf{Optimal transport in robot planning.}~While OT has several practical applications in problems of resource assignment and machine learning~\citep{peyre2019computational}, its application to robotics is scarce. Most applications consider swarm and multi-robot coordination~\citep{inoue2021optimal,krishnan2018distributed,bandyopadhyay2014probabilistic,kabir2021efficient,frederick2022collective}, while OT can be used for exploring while planning~\citep{kabir2020receding}, for imitation learning~\citep{haldar2023watch}, and for curriculum learning~\citep{klink2022curriculum}. A comprehensive review of OT in control is available in~\citep{chen2021optimal}. To the best of our knowledge, we are the first to introduce UOT-based inference of reactive robot planning and policy blending.

\vspace{-0.5cm}
\section{Conclusion}\label{sec:discussion}
We proposed an efficient hierarchical policy blending framework as entropic-regularized UOT, where the upper level evaluates the expert contributions based on the task costs and adapts the weight scaling of lower-level experts-agents. Our method introduces an efficient look-ahead evaluation into myopic composable frameworks, thereby improving their reactiveness, safety guarantees, and has better chances of avoiding local minima. There are multiple exciting directions to be explored, e.g., better exploration mechanisms in HiPBOT, as the current cost evaluation only allows as many rollouts as the number of experts. Further, casting the policy blending problem as OT opens a novel view about interactions of multi-experts to multi-agents, which is largely under-explored. 

\clearpage
\acks{This research is supported by the German Research Foundation through the collaborative projects METRIC4IMITATION (PE 2315/11-1) CHIRON, and the Emmy Noether Programme (CH 2676/1-1). The authors would like to thank Joao Carvalho for his feedback, and Snehal Jauhri for his advice on setting up the whole-body control environment.}
\bibliography{ebm, learning4mp, mp, ot, hipbi}

\end{document}